\documentclass[10pt,twocolumn,letterpaper]{article}

\usepackage{cvpr}              % To produce the CAMERA-READY version
\usepackage{graphicx}
\usepackage{amsmath}
\usepackage{amssymb}
\usepackage{booktabs}

\usepackage[pagebackref,breaklinks,colorlinks]{hyperref}

\usepackage[capitalize]{cleveref}
\crefname{section}{Sec.}{Secs.}
\Crefname{section}{Section}{Sections}
\Crefname{table}{Table}{Tables}
\crefname{table}{Tab.}{Tabs.}

\def\cvprPaperID{*****} % *** Enter the CVPR Paper ID here
\def\confName{CVPR}
\def\confYear{2023}

\begin{document}

\title{Synthetic Sample Selection for Generalized Zero-Shot Learning}

\author{Shreyank N Gowda\\
University of Edinburgh\\
{\tt\small s.narayana-gowda@sms.ed.ac.uk}
}
\maketitle

%%%%%%%%% ABSTRACT
\begin{abstract}
Generalized Zero-Shot Learning (GZSL) has emerged as a pivotal research domain in computer vision, owing to its capability to recognize objects that have not been seen during training. Despite the significant progress achieved by generative techniques in converting traditional GZSL to fully supervised learning, they tend to generate a large number of synthetic features that are often redundant, thereby increasing training time and decreasing accuracy. To address this issue, this paper proposes a novel approach for synthetic feature selection using reinforcement learning. In particular, we propose a transformer-based selector that is trained through proximal policy optimization (PPO) to select synthetic features based on the validation classification accuracy of the seen classes, which serves as a reward. The proposed method is model-agnostic and data-agnostic, making it applicable to both images and videos and versatile for diverse applications. Our experimental results demonstrate the superiority of our approach over existing feature-generating methods, yielding improved overall performance on multiple benchmarks.
\end{abstract}

%%%%%%%%% BODY TEXT
\section{Introduction}
\label{sec:intro}

In recent years, deep learning \cite{resnet,densenet,vgg,colornet} has made remarkable strides in recognition accuracy, approaching human levels. However, the practical implementation of deep learning models is limited by the need for a significant number of labeled samples and the high cost of large-scale datasets \cite{imagenet}. It is often infeasible to collect sufficient labeled data for all classes, presenting a significant challenge for traditional supervised learning methods. To address this challenge, several approaches have been proposed, including semi-supervised learning, transfer learning, and few-shot learning. Zero-shot learning (ZSL) \cite{palatucci2009zero} is a subset of these methods, which refers to tasks where there is no training data available for unseen classes and disjoint label sets between training and test data.

ZSL is a unique form of cross-modal retrieval learning that relies on knowledge transfer from known classes to unknown classes through attribute sharing. The most common ZSL techniques use an intermediate semantic representation, such as visual features or semantic word vectors \cite{socher2013zero,frome2013devise,ye2017zero,kodirov2017semantic}, shared between the labeled auxiliary dataset and the unlabeled target dataset. The projection from the low-level feature space to the semantic space is learned from the auxiliary dataset, without adapting to the target dataset. The Generalized ZSL scenario, where both seen and unseen class samples are available at test time, is considered to be more realistic than the standard ZSL setup \cite{xian2018zero}.

\begin{figure}[t]
    \centering
    \includegraphics[width=0.95\linewidth]{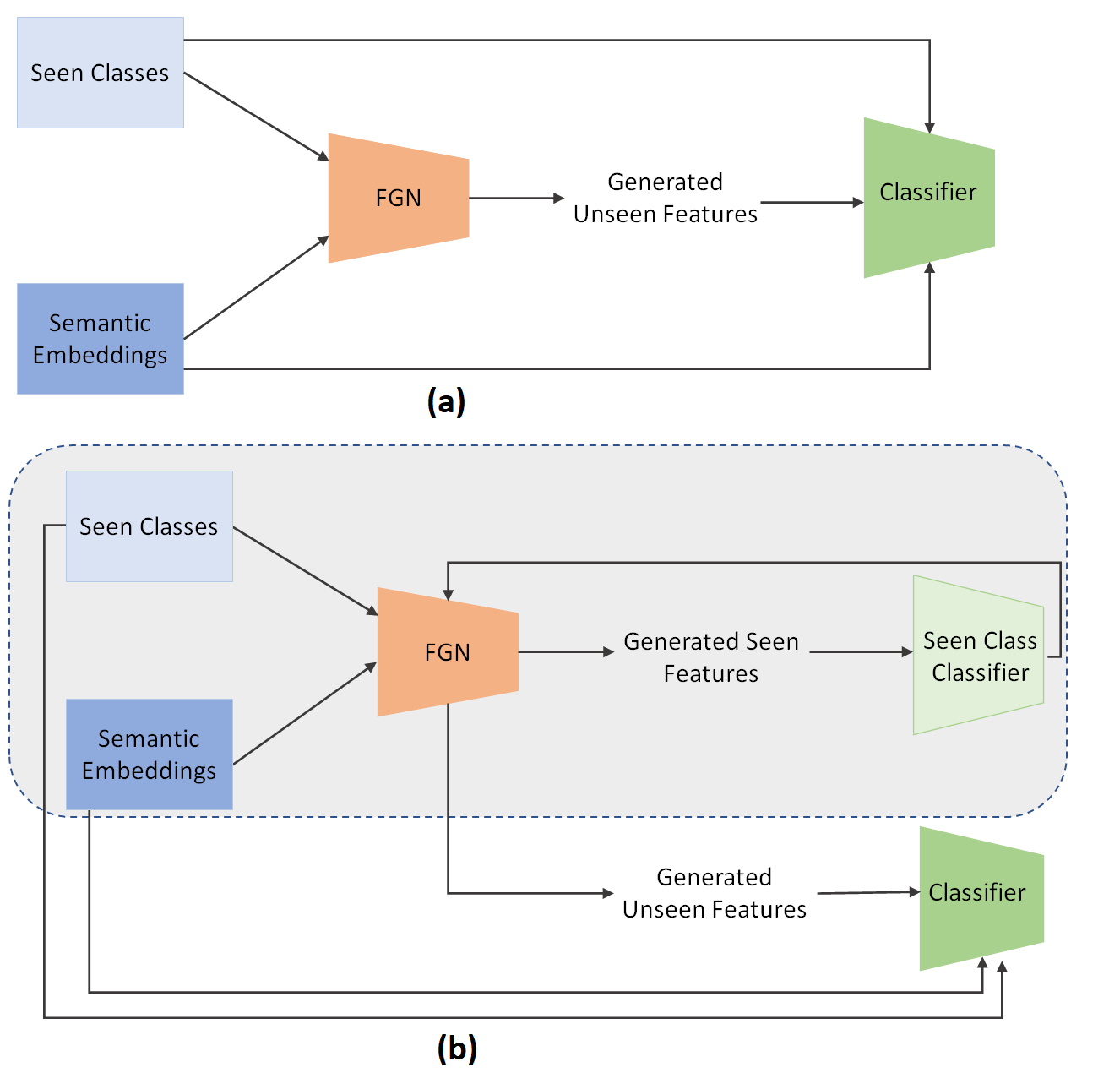}
    \caption{Comparison of feature-generating frameworks pipelines (a) standard pipeline where features that look ``real" are used to train the generator (b) proposed pipeline where the generator is trained based on the performance of the seen class classifier. }
    \label{fig:teaser}
\end{figure}

In Generalized ZSL, classifiers tend to be biased towards seen classes, leading to the misclassification of test samples from unseen classes as seen classes. To address the problem of the lack of visual data for unseen classes, researchers have proposed the use of Generative Adversarial Networks (GAN) \cite{goodfellow2020generative} to generate synthetic visual features by leveraging attribute databases. However, while GANs have helped in zero-shot learning \cite{verma2018generalized,clswgan,yu2020episode,cyclewgan}, they do not explicitly learn to represent data in a structured way that is easily interpretable by humans or other models. They also suffer from the problems of mode collapse \cite{jahaniansteerability}, class imbalance \cite{arora2017gans}, and computational expense \cite{guo2019deep} when generating high-dimensional data. But what we care about most is that a lot of synthetic samples generated are used directly for training a classifier without studying if these samples actually help the classifier learn. Instead, these samples are chosen based on ``realness". Figure~\ref{fig:teaser} shows a comparison of the standard pipeline and our proposed pipeline for feature-generating approaches.

To address the limitations of GANs in synthetic feature selection, we propose a novel reinforcement learning-based approach that automatically selects generated features that improve model performance. Specifically, we use a transformer model \cite{vaswani2017attention} for synthetic sample selection and use validation classification accuracy as the reward for RL training. We employ the proximal policy optimization (PPO) \cite{ppo} algorithm to update the model weights during training. Our proposed approach aims to pick samples that help classification and not just generate real-looking samples. We dub our synthetic sample selection method as ``\textbf{SPOT}" for \textbf{S}election using \textbf{P}roximal policy \textbf{O}p\textbf{T}imization.

Furthermore, our proposed approach is model-agnostic and data-agnostic, as we evaluate our method on multiple benchmark datasets in images and videos and various feature-generating models. Our comprehensive experiments demonstrate that our approach consistently improves model performance across different datasets and models, highlighting the effectiveness and versatility of our proposed method. By leveraging RL-based synthetic feature selection, we can more effectively generate synthetic data that captures the underlying structure of the data, improving the generalization performance of downstream models.

\section{Related Work}
\label{sec:background}

\paragraph{Zero-Shot Learning in Images}

Zero-shot learning (ZSL) is a challenging problem in computer vision, where the task is to recognize object categories without any training examples for them. Various approaches have been proposed to solve this problem in images. One of the early works \cite{lampert2009learning} in this field used attributes, such as color and shape, to describe the object categories and mapped them to a visual space. They then used a nearest-neighbor classifier to recognize unseen object categories. However, this approach suffers from the semantic gap problem, where the attributes do not always correlate well with the visual features.

To address this problem, more recent works have explored the use of deep learning techniques to learn a joint embedding space for the visual and semantic features. One such approach is proposed by Frome et al. (2013) \cite{frome2013devise}, where they used a deep neural network to learn a joint embedding space for the visual and textual features of the objects. They then used a nearest-neighbor classifier to recognize unseen object categories. Another approach is proposed by Socher et al. (2013) \cite{socher2013zero}, where they used a recursive neural network to learn a compositional representation of the textual descriptions of the object categories. 

More recently, there has been a trend towards using generative models to solve the ZSL problem. One such approach is proposed by Xian et al. \cite{clswgan}, where they used a generative adversarial network (GAN) to generate visual features for unseen object categories. They then used a joint embedding space to match the generated features with the semantic features and recognize unseen object categories. Another approach is proposed by Schonfeld et al. (2019) \cite{schonfeld2019generalized}, where they used a GAN to generate visual features conditioned on the textual descriptions of the object categories. ZeroGen~\cite{zerogen} uses pre-trained language models to synthesize a dataset for a zero-shot task and then trains a small task model on it. NereNet~\cite{nerenet} generates unseen samples by combining noise from similar seen classes with unseen class attributes using GAN. CMC-GAN~\cite{cmcnet}, performs data hallucination of unseen classes by performing semantics-guided intra-category knowledge transfer across image categories.

We add our proposed module of synthetic sample selection to multiple feature-generating frameworks and show that this leads to improved performance for all these models across all datasets.

\begin{figure*}[t]
    \centering
    \includegraphics[width=0.85\linewidth]{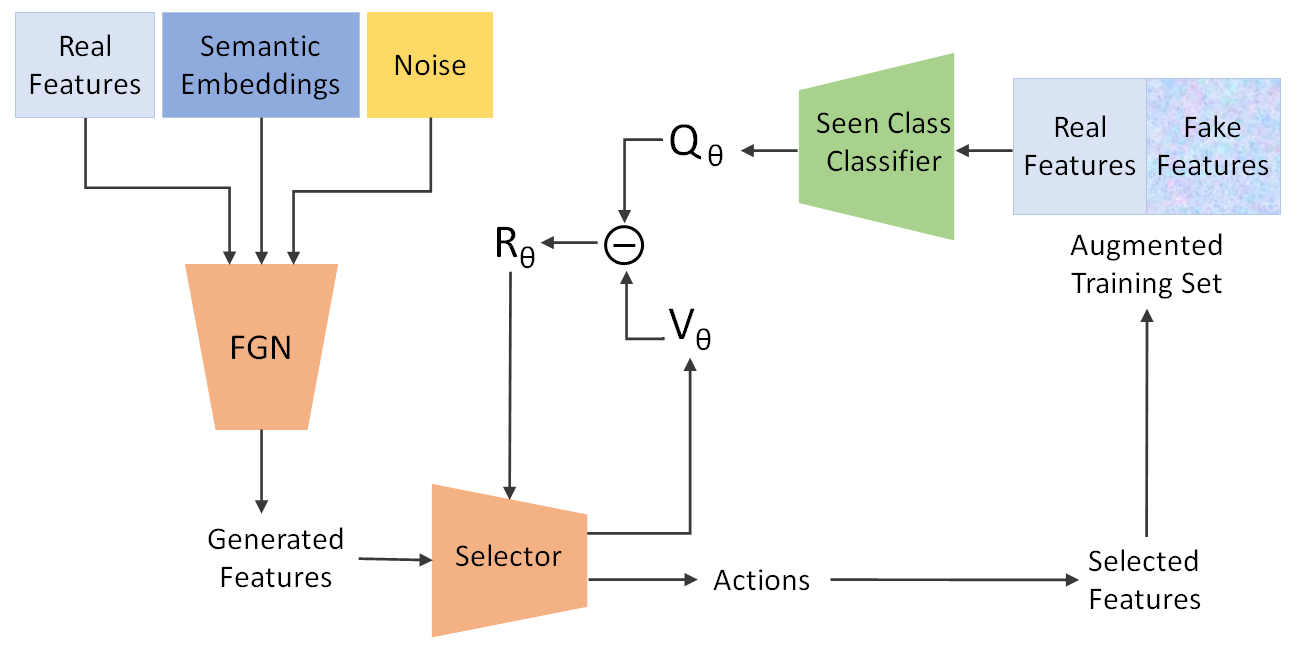}
    \caption{Overall pipeline of our proposed SPOT. The feature generator generates features that the selector module ranks based on the seen class classifier's performance. The selector is updated based on the performance of the classifier on the selected features. The proposed pipeline is model and data-agnostic.}
    \label{fig:overview}
\end{figure*}

\paragraph{Zero-Shot Learning in Videos}

The initial study by Rohrbach et al.\cite{rohrbach12eccv} utilized script data from cooking activities to facilitate the transfer of knowledge to unseen categories. Gan et al.\cite{gan2016learning} considered each action class as a domain and tackled the problem of identifying semantic representations as a multisource domain generalization task. To extract semantic embeddings of class labels, popular approaches employ label embeddings such as word2vec~\cite{word2vec}, which solely requires class names. Several methods have used a shared embedding space between video features and class labels~\cite{xu2016multi,xu2017transductive}, error-correcting codes \cite{qin2017zero}, pairwise relationships between classes \cite{gan2016concepts}, interclass relationships \cite{gan2015exploring}, out-of-distribution detectors~\cite{OD}, synthetic features~\cite{syn, GGM2018} and graph neural networks~\cite{gao2019know}.

Recently, it has been observed that clustering joint visual-semantic features results in better representations for zero-shot action recognition~\cite{claster}. Similar to CLASTER, ReST \cite{rest} jointly encodes video data and textual labels for zero-shot action recognition. In ReST, transformers are utilized to conduct modality-specific attention. On the other hand, JigSawNet \cite{jigsaw} models visual and textual features jointly but disassembles videos into atomic actions in an unsupervised manner and establishes group-to-group relationships between visual and semantic representations instead of the one-to-one relationships that CLASTER and ReST establish.

However, since we only evaluate on feature generating approaches, we compare directly to OD~\cite{OD}, CLSWGAN~\cite{clswgan}, GGM~\cite{GGM2018} and Bi-dir GAN~\cite{syn} and show that using our synthetic feature selection approach all methods can be improved significantly.

\paragraph{Reinforcement Learning for Data Valuation}

The quantification of data value for a specific machine learning task, known as data valuation \cite{smart,jia2019towards,wu2022davinz}, has numerous applications such as domain adaptation, corrupted sample detection, and robust learning. Various techniques have been proposed to estimate data values based on different criteria, including influence functions, Shapley values, leave-one-out errors, and data deletion. However, these methods are computationally expensive, necessitate model perturbations or retraining, and do not consider the interactions among data points. Recently, an adaptive approach to data valuation using reinforcement learning \cite{yoon2020data}, in which data values are jointly learned with the predictor model using a data value estimator that is trained using a reinforcement signal reflecting task performance. Similarly, Learn2Augment~\cite{L2A} performs data valuation of augmented samples created by combining foreground and background videos using reinforcement learning to quantify the value of an augmented sample.

Synthetic sample selection~\cite{ye2020synthetic} for medical image segmentation is an under-investigated research area that focuses on the quality control of synthetic images for data augmentation purposes. Synthetic images are not always realistic and may contain misleading features that distort data distribution when mixed with real images. As a result, the effectiveness of synthetic images in medical image recognition systems cannot be ensured when they are randomly added without quality assurance. A reinforcement learning-based synthetic sample selection approach is proposed in which synthetic images containing reliable and informative features are chosen.

However, none of the above approaches consider the extreme case of zero-shot learning. In the case of zero-shot learning, synthetic features are often biased towards seen classes, and the generated synthetic features do not represent the true distribution. Training a model to generate realistic-looking features will produce features similar to the training distribution without any guarantee on the effect on zero-shot evaluation. Therefore, we propose a data valuation method for synthetic features based on classification performance rather than their realism.

\section{Methodology}
\label{sec:method}

The overall framework of the proposed method is visually depicted in Figure~\ref{fig:overview}, which provides an illustrative overview of the various components employed. In this section, we delve deeper into the individual constituents of the model and explore in detail the novel SPOT selector that has been put forth. It is imperative to note that the proposed pipeline is model and data-agnostic, which implies that the choice of classifier model and network backbone is dependent solely on the feature-generating framework itself.

The development of the SPOT selector draws significant inspiration from the synthetic sample selector introduced in ~\cite{ye2020synthetic}. However, our approach seeks to tackle the more challenging task of zero-shot learning, where data from unseen classes is limited. Furthermore, we demonstrate that training the selector using data from seen classes can enhance the selection of better features for data from unseen classes. This approach represents a significant contribution and serves to bridge the gap between the seen and unseen class data.

\subsection{Feature Generating Network (FGN)}
As previously highlighted, it is worth noting that the proposed pipeline is entirely independent of the feature-generating approach itself. This aspect renders the framework highly versatile and adaptable to a diverse range of feature-generating models, which may be employed in place of the FGN utilized in this study.

Examples of alternate feature-generating models that could be employed include the WGAN~\cite{clswgan} and Cycle-WGAN~\cite{cyclewgan}. The utilization of such models would permit the proposed pipeline to be seamlessly integrated into a broader range of applications and extend the reach and scope of the framework. The flexibility afforded by this design decision represents a critical feature of the proposed pipeline and enables the framework to be readily adapted to a range of diverse use cases as shown with consistent improvements in multiple image and video datasets.

\subsection{Selector}

The reasoning behind selecting the particular selector is rooted in the interdependence of features among the potential images. We hypothesise that the sequence in which the features are generated is not completely autonomous, as the later additions must differentiate themselves from the earlier ones in order to ensure diversity across the entire set of augmented training data. We use a transformer-based architecture~\cite{vit} to be our selector. The input is a feature vector of a dimension dependent on the FGN used. The goal of the selector is to tell us if the generated feature vector is good for classification performance. The selector takes in a feature vector and outputs a score that tells us how good that feature vector is for classification. To do this, the selector outputs a binary action: select or not select. However, we do not have ground truth to tell us how good the generated feature is and hence optimizing the selector is not trivial.

To address the possibility of a relationship between augmented images without relying heavily on sequential assumptions, we utilize the self-attention mechanism through the implementation of the transformer~\cite{vit} model as our selector. The transformer architecture eliminates all recurrent structures, requiring feature vectors to be combined with their positional embeddings using sinusoidal functions prior to being input into the encoder layer of the transformer. The primary component of the transformer encoder is the multi-head attention block, comprised of $n$ self-attention layers, where $n$ denotes the number of heads. In each self-attention layer, input features are projected to three separate feature spaces - query $Q$, key $K$, and value $V$- by multiplying learnable weight matrices. The resulting attention map is obtained through the following process:

\begin{equation}
    Attention(Q, K, V) = softmax(\frac{QK^T}{\sqrt{d_k}})V
\end{equation}

Each head of the multi-head attention block represents a distinct projected feature space for the input, achieved by multiplying the same input embedding with different weight matrices. These separate outputs are then concatenated to form the final attention map, which is expressed as:

\begin{equation}
    MultiHead(Q, K, V) = Concat(head_1, ..., head_h)W^O
\end{equation}

\begin{equation}
    head_i = Attention(Q W^Q_i, K W^K_i, V W^V_i)
\end{equation}

Once we obtain the attention map, the context vector is then fed to the feed-forward layer as follows:
\begin{equation}
    F(x) = max(((0, xW_{1} + b_{1})W_{2} + b_{2})W_{3}+b_{3}...)W_{n}+b_{n}
\end{equation}

Given that the objective of the selector is to produce a binary action for every input feature vector, the decoder for the transformer model is a linear layer that functions as the policy network. Overall, the use of the transformer as the selector within our image selection framework, based on reinforcement learning, is beneficial due to its self-attention mechanism that effectively captures the interdependencies among the input feature vectors. We conducted a thorough ablation study regarding the selection of the selector and this can be seen in Section 4.2.

To optimize the selector, we turn to reinforcement learning as this is a common solution ~\cite{yoon2020data,L2A}. In particular, we use proximal policy optimization~\cite{ppo}. Details are explained in the next section.

\subsection{Proximal Policy Optimization}

As previously stated, we turn to reinforcement learning approach to update the selector model. A proficient policy gradient method is fundamental to effectively utilize reward feedback as input to the selector in the reinforcement learning process. Among various policy gradient algorithms, Proximal Policy Optimization (PPO)~\cite{ppo} has gained popularity due to its computational efficiency and satisfactory performance, surpassing previous approaches like TRPO~\cite{trpo}. Additionally, PPO alleviates the instability encountered during RL training. PPO achieves comparable performance with reduced complexity by replacing the KL convergence constraint enforced in TRPO with a clipped probability ratio between the current and previous policy within a small interval around 1. At each time step t, with $A_{\theta}$ representing the advantage function, the objective function is defined as follows:

\begin{multline}
      L(\theta) = E [min(\gamma_{\theta}(t)A_{\theta}(s_{t}, a_{t}), \\ clip(\gamma_{\theta}(t),1-\epsilon, 1 + \epsilon)A_{\theta}(s_{t}, a_{t}))] 
\end{multline}

Here, $A_{\theta}(s_{t}, a_{t}) = Q_{\theta}(s_{t}, a_{t}) - V_{\theta}(s_{t}, a_{t})$. As a component of the transformer output, the learned state-value $V_{\theta}(s_{t}, a_{t})$  serves as a baseline for the q-value to mitigate the variance of rewards during the training process. The probability of actions is denoted by $\pi$. The q-value at time t, $Q_{\theta}(s_{t}, a_{t})$, is defined as a smoothed version of the maximum validation accuracy observed among the last five epochs in the classification task. As our target tasks are trained on the seen class data and needs to generalize to the unseen class data, it is crucial to obtain a robust estimation of the reward's changing pattern. To achieve this, we employ the Exponential Moving Average (EMA) algorithm to smooth the original reward curve. Thus, the final reward at time $t$ is obtained as follows:

\begin{equation}
\hat{Q}_\theta\left(s_t, a_t\right)=\left\{\begin{array}{cc}
\mathrm{Q}_\theta\left(s_t, a_t\right), & t=1 \\
\alpha \hat{Q}_\theta\left(s_{t-1}, a_{t-1}\right)+ \\ (1-\alpha) \mathrm{Q}_\theta\left(s_t, a_t\right), & t>1
\end{array} .\right.
\end{equation}

Drawing inspiration from the concept of importance sampling, the weight assigned to the current policy is influenced by earlier policies. The probability ratio between the previous and current policies, denoted by $\gamma_{\theta}(t)$, is mathematically defined as:
\begin{equation}
\gamma_\theta(t)=\frac{\pi_\theta\left(a_t \mid s_t\right)}{\pi_\theta\left(a_{t-1} \mid s_{t-1}\right)}
\end{equation}

Here, $a_t \in \mathbb{R}^{N \times 2}$ refers to the number of synthetic samples in the candidate pool. If at any given timestep $t$, $a_{i}(t)=0$ then $i$ is discarded. Else, it is added to the original training set.

\section{Experimental Analysis}
\label{sec:exp}

\subsection{Implementation Details}

Since we propose a plug-and-play component to feature-generating networks, the backbone and technical details follow the exact same implementation that the feature-generating model uses. Here, we talk about the technical details of running SPOT.

The candidate features generated by the feature-generating framework are passed into the selector network which is a 8-layer encoder having a 8-head multi-head attention block. The output is a vector $A_{\theta}$ which is the action vector and a value vector $V_{\theta}$ that is used together to calculate the reward for the policy gradient algorithm.

The classifier network depends on the feature-generating network being used since our proposed method is model agnostic. Similar to ~\cite{L2A,ye2020synthetic} we use the EMA-smoothed validation accuracy obtained from the last 5 epochs as the reward with $\alpha$=0.5 when using the classifier on the validation set. As long as this average is increasing, we continue updating our policy. The policy function $\pi$ is obtained from the softmax layer of the seen-class classifier model.

$\epsilon$ in Eq. 4 is set to 0.15 (see Ablations for empirical comparison), this helps to set an upper and lower bound at the current time step $t$ and previous one $t-1$ for the ratio of the policy function. The number of selected synthetic features are dependent on the selector and varies according to model and dataset. However, we set the learning rate to be fixed for the PPO at 2e - 04. 

\subsection{Ablation Study}

\begin{figure}[t]
    \centering
    \includegraphics[width=0.9\linewidth]{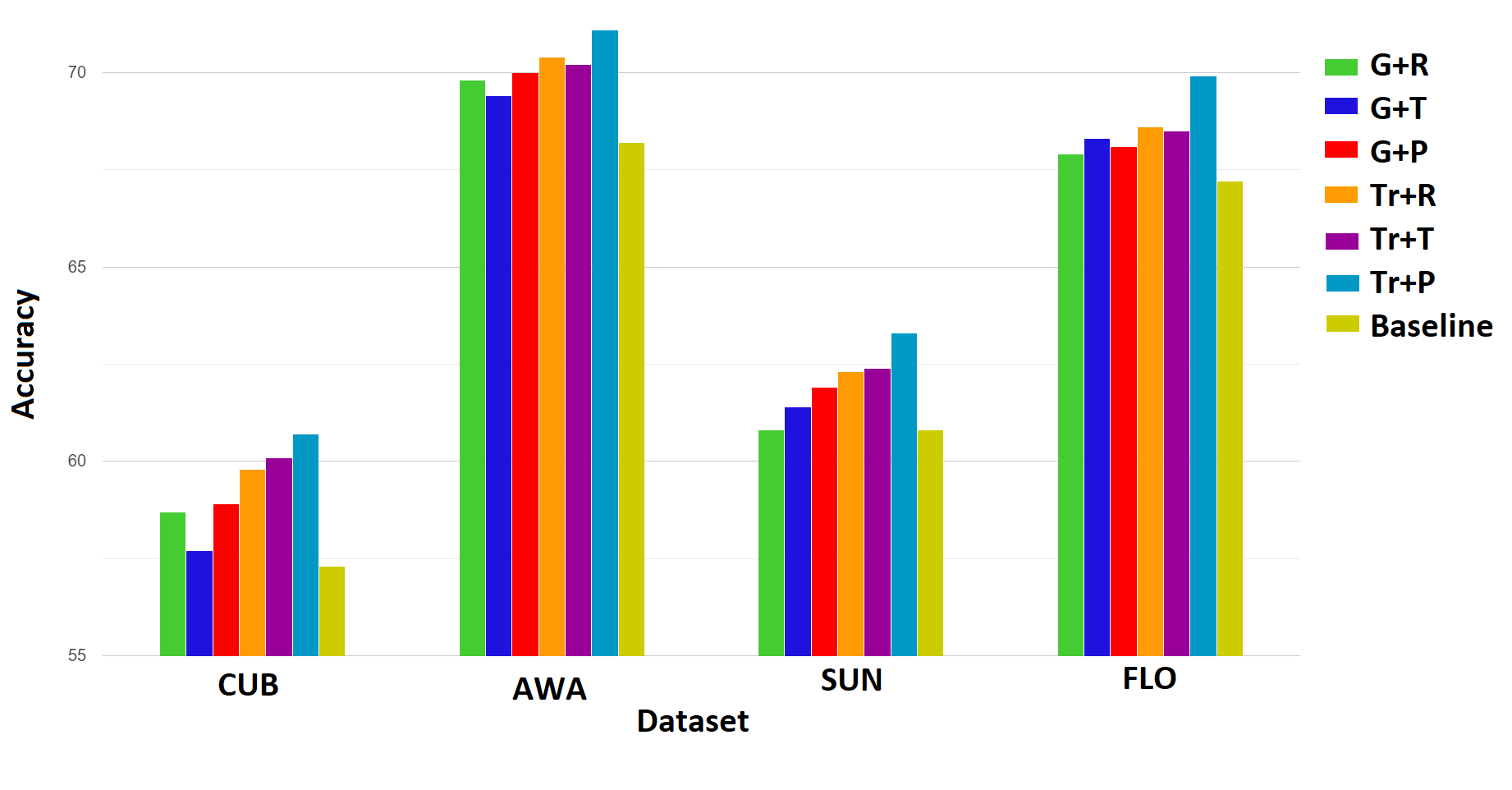}
    \caption{Ablation comparison of different combinations of RL algorithms with selector choices (GRU or Transformer). `G' = GRU, `Tr' = Transformer, `R' = REINFORCE, `T' = TRPO and `P' = PPO.}
    \label{fig:ablation}
\end{figure}

We have made a few choices with regards to the hyperparameter selection and choice of RL optimization algorithms and in this section, we show empirical reasons why the choices were made. Figure~\ref{fig:ablation} shows the performance differences when using different RL optimization algorithms to modify the selector. Similar to ~\cite{ye2020synthetic}, we also consider alternative choices such as GRU and LSTMs for the selector. In terms of RL algorithms, we compare to REINFORCE~\cite{reinforce} and TRPO~\cite{trpo}.

\subsection{Images}

\setlength{\tabcolsep}{2pt}
\begin{table}[t]
\small
\begin{center}
\begin{tabular}{|l|l|l|l|l|}
\hline
Model                        & CUB  & AWA & SUN  & FLO  \\
\hline \hline
WGAN                         & 57.3 & 68.2 & 60.8 & 67.2 \\
WGAN + \textbf{SPOT}         & \textbf{60.7} & \textbf{71.1} & \textbf{63.3} & \textbf{69.9} \\
\hline
Cycle-WGAN                   & 57.8 & 65.6 & 59.7 & 68.6 \\
Cycle-WGAN + \textbf{SPOT} & \textbf{61.1} & \textbf{69.7} & \textbf{62.5} & \textbf{70.9} \\
\hline
f-VAEGAN & 61.0 & 71.1 & 64.7 & 67.7 \\
f-VAEGAN + \textbf{SPOT} & \textbf{62.8} & \textbf{72.7} & \textbf{66.0} & \textbf{69.2} \\
\hline
CMC-GAN                      & 61.4 & 71.4 & 63.7 & 69.8 \\
CMC-GAN + \textbf{SPOT} & \textbf{62.9} & \textbf{73.1} & \textbf{65.1} & \textbf{71.9} \\
\hline
\end{tabular}
\end{center}
\caption{Results on zero-shot image classification using recent feature-generating frameworks.
}
\label{tbl:results:zsl_image}
\end{table}

\begin{table*}[htb]
\small
\begin{center}
\begin{tabular}{|l|c|c|c|c|c|c|c|c|c|c|c|c|}
\hline
\multicolumn{1}{|l|}{Model} & \multicolumn{3}{c|}{CUB}                                                                                  & \multicolumn{3}{c|}{AWA1}                                                                                 & \multicolumn{3}{c|}{SUN}                                                                                  & \multicolumn{3}{c|}{FLO}                                                                                  \\ \cline{2-13} 
\multicolumn{1}{|l|}{}                       & \multicolumn{1}{c|}{S}            & \multicolumn{1}{c|}{U}            & \multicolumn{1}{c|}{H}            & \multicolumn{1}{c|}{S}            & \multicolumn{1}{c|}{U}            & \multicolumn{1}{c|}{H}            & \multicolumn{1}{c|}{S}            & \multicolumn{1}{c|}{U}            & \multicolumn{1}{c|}{H}            & \multicolumn{1}{c|}{S}            & \multicolumn{1}{c|}{U}            & \multicolumn{1}{c|}{H}            \\ \hline
WGAN                                         & 43.7                              & 57.7                              & 49.7                              & 57.9                              & 61.4                              & 59.6                              & 42.6                              & 36.6                              & 39.4                              & 59.0                              & 73.8                              & 65.6                              \\
WGAN+\textbf{SPOT}                                    & \textbf{44.1}                     & \textbf{60.9}                     & \textbf{51.1}                     & \textbf{58.6}                     & \textbf{64.9}                     & \textbf{61.6}                     & \textbf{42.8}                     & \textbf{39.1}                     & \textbf{40.9}                     & \textbf{59.3}                     & \textbf{75.9}                     & \textbf{66.6}                     \\
\hline
Cycle-WGAN                                   & 46.0                              & 60.3                              & 52.2                              & 56.4                              & 63.5                              & 59.7                              & \textbf{48.3}                     & 33.1                              & 39.2                              & 59.1                              & 71.1                              & 64.5                              \\
Cycle-WGAN+\textbf{SPOT}                              & \textbf{46.5}                     & \textbf{62.9}                     & \textbf{53.5}                     & \textbf{56.9}                     & \textbf{66.1}                     & \textbf{61.1}                     & 48.1                              & \textbf{36.2}                     & \textbf{41.3}                     & \textbf{59.4}                     & \textbf{74.4}                     & \textbf{66.1}                     \\
\hline
f-VAEGAN                                     & 48.4                              & 60.1                              & 53.6                              & 57.6                              & 70.6                              & 63.5                              & 45.1                              & 38.0                              & 41.3                              & 56.8                              & 74.9                              & 64.6                              \\
f-VAEGAN+\textbf{SPOT}                               & \textbf{48.8}                     & \textbf{62.8}                     & \textbf{54.9}                     & \textbf{57.9}                     & \textbf{73.3}                     & \textbf{64.7}                     & \textbf{45.5}                     & \textbf{41.1}                     & \textbf{43.2}                     & \textbf{57.0}                     & \textbf{77.2}                     & \textbf{65.6}                     \\
\hline
CMC-GAN                                      & 52.6                              & 65.1                              & 58.2                              & 63.2                              & 70.6                              & 66.7                              & 48.2                              & 40.8                              & 44.2                              & 64.5                              & 80.2                              & 71.5                              \\
CMC-GAN+\textbf{SPOT}                                 & \textbf{53.1}                     & \textbf{66.7}                     & \textbf{59.1}                     & \textbf{63.3}                     & \textbf{73.8}                     & \textbf{68.1}                     & \textbf{48.9}                     & \textbf{44.1}                     & \textbf{46.4}                     & \textbf{64.6}                     & \textbf{82.8}                     & \textbf{72.6}                     \\
\hline
NereNET                                      & 51.0                              & 56.5                              & 53.6                              & -                                 & -                                 & -                                 & 45.7                              & 38.1                              & 41.6                              & -                                 & -                                 & -                                 \\
NereNET+\textbf{SPOT}                                 & \textbf{51.3}                     & \textbf{58.4}                     & \textbf{54.6}                     & -                                 & -                                 & -                                 & \textbf{45.9}                     & \textbf{40.4}                     & \textbf{43.0}                     & -                                 & -                                 & -                                 \\
\hline
FREE                                         & \textbf{55.7} & 59.9    & 57.7         & 62.9         & 69.4         & 66.0      & 47.4        & 37.2          & 41.7          & 67.4       & 84.5         & 75.0     \\
FREE+\textbf{SPOT}                                    & 55.5         & \textbf{62.2} & \textbf{58.6} & \textbf{63.1} & \textbf{72.1} & \textbf{67.3} & \textbf{47.8} & \textbf{39.9} & \textbf{43.5} & \textbf{67.8} & \textbf{86.3} & \textbf{75.9} \\
\hline
DAA                                          & 66.1          & 65.5 & 65.8        & 64.3         & 76.6        & 69.9         & 47.8          & 38.7        & 42.8         & -                                 & -                                 & -                                 \\
DAA+\textbf{SPOT}                                     & \textbf{66.3} & \textbf{67.7} & \textbf{67.0} & \textbf{64.6} & \textbf{77.9} & \textbf{70.6} & \textbf{48.1} & \textbf{40.3} & \textbf{43.8} & -                                 & -                                 & -                               \\ 
\hline
\end{tabular}
\end{center}
\caption{Results on generalized zero-shot image classification on 4 challenging benchmarks.
}
\end{table*}

\subsubsection{Datasets and Evaluation Protocol}

Our method is evaluated on four challenging benchmark datasets, namely AWA \cite{lampert2013attribute}, CUB (Caltech UCSD Birds 200) \cite{cub}, SUN (SUN Attribute) \cite{sun}, and FLO \cite{nilsback2008automated}. CUB and SUN are fine-grained datasets, while AWA and FLO are coarse-grained datasets.  We adopt the same seen/unseen splits and class embeddings to ensure consistency with previous work as in \cite{xian2017zero}. AWA1 contains 30,475 instances across 50 categories. CUB comprises 11,788 images of 200 bird classes (150/50 for seen/unseen classes) with 312 attributes. SUN contains 14,340 images from 717 scene classes (645/72 for seen/unseen classes) with 102 attributes. FLO consists of 8,189 images from 102 flower classes with an 82/20 class split for seen and unseen classes, respectively. These datasets are widely used in the literature, enabling a direct comparison of our results with those of previous studies.

\subsubsection{Zero-Shot Learning}

We compare strictly with recent state-of-the-art feature-generating approaches and as such compare to WGAN~\cite{clswgan}, Cycle-WGAN~\cite{cyclewgan}, f-VAEGAN~\cite{fvaegan}, NereNet~\cite{nerenet} and CMC-GAN~\cite{cmcnet}. Table~\ref{tbl:results:zsl_image} shows the results. We see consistent gains with increase of up to 4.1\% when we add the proposed SPOT to any of the models.

\subsubsection{Generalized Zero-Shot Learning}

We perform a much more extensive comparison in the generalized setting as this is where most feature generating frameworks perform experiments. We compare against WGAN~\cite{clswgan}, Cycle-WGAN~\cite{cyclewgan}, f-VAEGAN~\cite{fvaegan}, CMC-GAN~\cite{cmcnet}, NereNet~\cite{nerenet}, FREE~\cite{free} and DAA~\cite{daa}. We see consistent improvements on the unseen class accuracies as this is where selected features make a difference. We see gains of up to 3.3\% on the unseen class accuracies. As a result, there is consistent improvement on the harmonic mean of the seen and unseen class accuracies as well.

\subsection{Videos}

\subsubsection{Datasets and Evaluation Protocol}

For videos, we use the widely adopted Olympic Sports \cite{olympics}, HMDB-51 \cite{hmdb}, and UCF-101 \cite{ucf101} datasets to evaluate our method for zero-shot action recognition and compare it against recent state-of-the-art feature generating models \cite{OD, clswgan, finegrain}. The aforementioned datasets comprise 783, 6766, and 13320 videos and are associated with 16, 51, and 101 classes, respectively. To enable comparison with existing works \cite{OD, clswgan, finegrain, GGM2018, syn}, we adopt the widely used 50/50 splits proposed by Xu et al. \cite{xu2017transductive}, where half of the classes are considered as seen and the other half as unseen. We report the average accuracy and standard deviation over 10 independent runs, following previous approaches.

Moreover, we extend our evaluation to include TruZe~\cite{truze}, which was recently introduced to address the issue of overlapping classes between the pre-training dataset (Kinetics~\cite{i3d}) and the unseen classes in zero-shot settings. The TruZe split acknowledges the presence of such overlapping classes, which contradicts the fundamental assumption that the unseen classes have not been previously seen.

\subsubsection{Zero-Shot Learning}

Table~\ref{tbl:results:zsl} shows the effect of using our proposed SPOT selector to enhance the performance of state-of-the-art feature-generating frameworks on the zero-shot setting. We compare with the most recent best-performing methods, which include the Bi-Dir GAN~\cite{syn}, GGM~\cite{GGM2018}, OD~\cite{OD}, WGAN~\cite{clswgan} and FFG~\cite{finegrain} (fine-grained feature generation framework).

\setlength{\tabcolsep}{2pt}
\begin{table}[t]
\small
\begin{center}
\begin{tabular}{|l|c|c|c|}
\hline
Method & Olympics & HMDB51 & UCF101\\
\hline\hline
Bi-Dir GAN \cite{syn} & 53.2 $\pm$ 10.5 & 21.3 $\pm$ 3.2 &	24.7 $\pm$ 3.7\\
Bi-Dir GAN \cite{syn} + \textbf{SPOT} & \textbf{56.6 $\pm$ 10.1} & \textbf{25.1 $\pm$ 3.4} &	\textbf{27.7 $\pm$ 3.5}\\
\hline

GGM \cite{GGM2018} & 57.9 $\pm$ 14.1 & 20.7 $\pm$ 3.1 & 24.5 $\pm$ 2.9 \\
GGM \cite{GGM2018} + \textbf{SPOT} & \textbf{62.4 $\pm$ 12.4} & \textbf{25.1 $\pm$ 2.8} & \textbf{27.4 $\pm$ 2.5} \\
\hline
OD \cite{OD} & 65.9 $\pm$ 8.1 & 30.2 $\pm$ 2.7 & 38.3 $\pm$ 3.0\\
OD \cite{OD} + \textbf{SPOT} & \textbf{68.7 $\pm$ 7.5} & \textbf{34.4 $\pm$ 2.2} & \textbf{40.9 $\pm$ 2.6}\\
\hline
WGAN \cite{clswgan} & 64.7 $\pm$ 7.5 & 29.1 $\pm$ 3.8 & 37.5 $\pm$ 3.1 \\
WGAN \cite{clswgan} + \textbf{SPOT} & \textbf{68.1 $\pm$ 7.1} & \textbf{33.8 $\pm$ 2.4} & \textbf{40.6 $\pm$ 2.4} \\
\hline

FFG \cite{finegrain} & - & 32.4 $\pm$ 2.3 & 27.6 $\pm$ 2.4 \\
FFG \cite{finegrain} + \textbf{SPOT} & - & \textbf{35.9 $\pm$ 2.5} & \textbf{30.9 $\pm$ 2.2} \\
\hline
\end{tabular}
\end{center}
\caption{Results on zero-shot action recognition on the Olympics, HMDB51 and UCF101 datasets. %* run by us with author's code on same splits as ours. %\ls{centering is off} \ls{we should flip the rows for OD and WGAN to make it easier to compare with the rows above}
}
\label{tbl:results:zsl}
\end{table}

We observe that the proposed method consistently outperforms all approaches across all datasets by gains of up to 4.5\%.

\subsubsection{Generalized Zero-Shot Learning}

We evaluate SPOT on the generalized setting where at test time both seen and unseen class samples are used. Table~\ref{tbl:results:gzsl} shows the results, with the harmonic mean of the seen and unseen class accuracies. The proposed SPOT selector consistently improves all approaches across all datasets by gains of up to 4.2\%. 

\setlength{\tabcolsep}{2pt}
\begin{table}
\small
\begin{center}
\begin{tabular}{|l|c|c|c|}
\hline
Method & Olympics & HMDB51 & UCF101\\
\hline\hline
Bi-Dir GAN \cite{syn} & 44.2 $\pm$ 11.2 & 7.5 $\pm$ 2.4 &	22.7 $\pm$ 2.5\\
Bi-Dir GAN \cite{syn} + \textbf{SPOT} & \textbf{48.4 $\pm$ 10.3} & \textbf{10.9 $\pm$ 3.1} &	\textbf{25.1 $\pm$ 3.8}\\
\hline

GGM \cite{GGM2018} & 52.4 $\pm$ 12.2  & 20.1 $\pm$ 2.1 & 23.7 $\pm$ 1.2 \\
GGM \cite{GGM2018} + \textbf{SPOT} & \textbf{55.3 $\pm$ 11.9} & \textbf{23.4 $\pm$ 2.3} & \textbf{27.1 $\pm$ 3.1} \\
\hline

WGAN \cite{clswgan} & 59.9 $\pm$ 5.3 & 32.7 $\pm$ 3.4 & 44.4 $\pm$ 3.0 \\
WGAN \cite{clswgan} + \textbf{SPOT} & \textbf{62.4 $\pm$ 5.5} & \textbf{34.4 $\pm$ 2.9} & \textbf{46.2 $\pm$ 2.6} \\
\hline

OD\cite{OD}  & 66.2 $\pm$ 6.3 & 36.1 $\pm$ 2.2 & 49.4 $\pm$ 2.4\\
OD \cite{OD} + \textbf{SPOT} & \textbf{69.1 $\pm$ 6.5} & \textbf{38.2 $\pm$ 2.5} & \textbf{51.8 $\pm$ 2.5}\\
\hline

FFG \cite{finegrain} & - & 37.4 $\pm$ 1.9 & 40.4 $\pm$ 2.2 \\
FFG \cite{finegrain} + \textbf{SPOT} & - & \textbf{39.8 $\pm$ 1.4} & \textbf{42.8 $\pm$ 1.7} \\

\hline
\end{tabular}
\end{center}
\caption{Results on generalized zero-shot setting. Reported results are the harmonic mean of the seen and unseen class accuracies. 
}
\label{tbl:results:gzsl}
\end{table}

\subsection{Results on TruZe}

We also evaluate on the stricter TruZe~\cite{truze} split that ensures no overlap between the pre-trained model and test classes. Results are shown in Table~\ref{tbl:truze}. We only evaluate on OD and WGAN as these are the two feature-generating approaches that have results reported on the TruZe split. Again, we see that using SPOT for selection consistently improves the performance of the feature-generating framework.

\begin{table}[t]
\small
\begin{center}
\begin{tabular}{| *{5}{c|} }
\hline
Method & \multicolumn{2}{c|}{UCF101 } & \multicolumn{2}{c|}{HMDB51}\\
  & ZSL & GZSL  & ZSL & GZSL \\
 \hline\hline
WGAN  & 22.5 & 36.3  & 21.1 & 31.8 \\
WGAN + \textbf{SPOT} & \textbf{25.3} & \textbf{39.1} & \textbf{23.8} & \textbf{33.3} \\
\hline
OD & 22.9 & 42.4  & 21.7 & 35.5 \\
OD + \textbf{SPOT}  & \textbf{25.5} & \textbf{44.1} & \textbf{24.0} & \textbf{37.1} \\
\hline
\end{tabular}
\end{center}
\caption{Results on TruZe. We report the mean class accuracy for zero-shot; for generalized zero-shot, we report the harmonic mean of seen and unseen class accuracies. }
\label{tbl:truze}
\end{table}

\section{Conclusion}
\label{sec:conclusion}

In conclusion, although generative techniques have made significant progress in transforming traditional GZSL to fully supervised learning, they often generate redundant synthetic features, which can lead to reduced accuracy. To overcome this limitation, we have proposed an approach for synthetic feature selection using reinforcement learning, which involves training a transformer-based selector using proximal policy optimization (PPO) to select synthetic features based on the validation classification accuracy of seen classes as the reward. Our proposed method is model-agnostic and data-agnostic and hence is suitable for images and videos. The experimental results of our approach demonstrate its superiority over existing feature-generating methods, with improved overall performance observed across multiple benchmarks. Overall, our approach represents a significant contribution towards addressing the issue of synthetic feature redundancy in GZSL, and we believe that it has the potential to be widely applied in real-world scenarios.

%%%%%%%%% REFERENCES
{\small
\bibliographystyle{ieee_fullname}
\bibliography{PaperForReview}

\begin{thebibliography}{10}\itemsep=-1pt

\bibitem{arora2017gans}
Sanjeev Arora and Yi Zhang.
\newblock Do gans actually learn the distribution? an empirical study.
\newblock {\em arXiv preprint arXiv:1706.08224}, 2017.

\bibitem{i3d}
J. Carreira and Andrew Zisserman.
\newblock Quo vadis, action recognition? a new model and the kinetics dataset.
\newblock In {\em CVPR}, 2017.

\bibitem{free}
Shiming Chen, Wenjie Wang, Beihao Xia, Qinmu Peng, Xinge You, Feng Zheng, and
  Ling Shao.
\newblock Free: Feature refinement for generalized zero-shot learning.
\newblock In {\em Proceedings of the IEEE/CVF international conference on
  computer vision}, pages 122--131, 2021.

\bibitem{imagenet}
Jia Deng, Wei Dong, Richard Socher, Li-Jia Li, Kai Li, and Li Fei-Fei.
\newblock Imagenet: A large-scale hierarchical image database.
\newblock In {\em 2009 IEEE conference on computer vision and pattern
  recognition}, pages 248--255. Ieee, 2009.

\bibitem{vit}
Alexey Dosovitskiy, Lucas Beyer, Alexander Kolesnikov, Dirk Weissenborn,
  Xiaohua Zhai, Thomas Unterthiner, Mostafa Dehghani, Matthias Minderer, Georg
  Heigold, Sylvain Gelly, et~al.
\newblock An image is worth 16x16 words: Transformers for image recognition at
  scale.
\newblock {\em arXiv preprint arXiv:2010.11929}, 2020.

\bibitem{cyclewgan}
Rafael Felix, Ian Reid, Gustavo Carneiro, et~al.
\newblock Multi-modal cycle-consistent generalized zero-shot learning.
\newblock In {\em Proceedings of the European Conference on Computer Vision
  (ECCV)}, pages 21--37, 2018.

\bibitem{frome2013devise}
Andrea Frome, Greg~S Corrado, Jon Shlens, Samy Bengio, Jeff Dean, Marc'Aurelio
  Ranzato, and Tomas Mikolov.
\newblock Devise: A deep visual-semantic embedding model.
\newblock {\em Advances in neural information processing systems}, 26, 2013.

\bibitem{gan2016concepts}
Chuang Gan, Ming Lin, Yi Yang, Gerard De~Melo, and Alexander~G Hauptmann.
\newblock Concepts not alone: Exploring pairwise relationships for zero-shot
  video activity recognition.
\newblock In {\em Thirtieth AAAI conference on artificial intelligence}, 2016.

\bibitem{gan2015exploring}
Chuang Gan, Ming Lin, Yi Yang, Yueting Zhuang, and Alexander~G Hauptmann.
\newblock Exploring semantic inter-class relationships (sir) for zero-shot
  action recognition.
\newblock In {\em Proceedings of the National Conference on Artificial
  Intelligence}, 2015.

\bibitem{gan2016learning}
Chuang Gan, Tianbao Yang, and Boqing Gong.
\newblock Learning attributes equals multi-source domain generalization.
\newblock In {\em Proceedings of the IEEE conference on computer vision and
  pattern recognition}, pages 87--97, 2016.

\bibitem{gao2019know}
Junyu Gao, Tianzhu Zhang, and Changsheng Xu.
\newblock I know the relationships: Zero-shot action recognition via two-stream
  graph convolutional networks and knowledge graphs.
\newblock In {\em Proceedings of the AAAI Conference on Artificial
  Intelligence}, volume~33, pages 8303--8311, 2019.

\bibitem{goodfellow2020generative}
Ian Goodfellow, Jean Pouget-Abadie, Mehdi Mirza, Bing Xu, David Warde-Farley,
  Sherjil Ozair, Aaron Courville, and Yoshua Bengio.
\newblock Generative adversarial networks.
\newblock {\em Communications of the ACM}, 63(11):139--144, 2020.

\bibitem{L2A}
Shreyank~N Gowda, Marcus Rohrbach, Frank Keller, and Laura Sevilla-Lara.
\newblock Learn2augment: Learning to composite videos for data augmentation in
  action recognition.
\newblock In {\em European Conference on Computer Vision}, pages 242--259.
  Springer, 2022.

\bibitem{smart}
Shreyank~N Gowda, Marcus Rohrbach, and Laura Sevilla-Lara.
\newblock Smart frame selection for action recognition.
\newblock In {\em Proceedings of the AAAI Conference on Artificial
  Intelligence}, volume~35, pages 1451--1459, 2021.

\bibitem{claster}
Shreyank~N Gowda, Laura Sevilla-Lara, Frank Keller, and Marcus Rohrbach.
\newblock Claster: clustering with reinforcement learning for zero-shot action
  recognition.
\newblock In {\em European Conference on Computer Vision}, pages 187--203.
  Springer, 2022.

\bibitem{truze}
Shreyank~N Gowda, Laura Sevilla-Lara, Kiyoon Kim, Frank Keller, and Marcus
  Rohrbach.
\newblock A new split for evaluating true zero-shot action recognition.
\newblock {\em arXiv preprint arXiv:2107.13029}, 2021.

\bibitem{colornet}
Shreyank~N Gowda and Chun Yuan.
\newblock Colornet: Investigating the importance of color spaces for image
  classification.
\newblock In {\em Computer Vision--ACCV 2018: 14th Asian Conference on Computer
  Vision, Perth, Australia, December 2--6, 2018, Revised Selected Papers, Part
  IV 14}, pages 581--596. Springer, 2019.

\bibitem{guo2019deep}
Wenzhong Guo, Jianwen Wang, and Shiping Wang.
\newblock Deep multimodal representation learning: A survey.
\newblock {\em IEEE Access}, 7:63373--63394, 2019.

\bibitem{resnet}
Kaiming He, Xiangyu Zhang, Shaoqing Ren, and Jian Sun.
\newblock Deep residual learning for image recognition.
\newblock In {\em Proceedings of the IEEE conference on computer vision and
  pattern recognition}, pages 770--778, 2016.

\bibitem{finegrain}
Mingyao Hong, Xinfeng Zhang, Guorong Li, and Qingming Huang.
\newblock Fine-grained feature generation for generalized zero-shot video
  classification.
\newblock {\em IEEE Transactions on Image Processing}, 2023.

\bibitem{densenet}
Gao Huang, Zhuang Liu, Laurens Van Der~Maaten, and Kilian~Q Weinberger.
\newblock Densely connected convolutional networks.
\newblock In {\em Proceedings of the IEEE conference on computer vision and
  pattern recognition}, pages 4700--4708, 2017.

\bibitem{jahaniansteerability}
Ali Jahanian, Lucy Chai, and Phillip Isola.
\newblock On the" steerability" of generative adversarial networks.
\newblock In {\em International Conference on Learning Representations}.

\bibitem{jia2019towards}
Ruoxi Jia, David Dao, Boxin Wang, Frances~Ann Hubis, Nick Hynes, Nezihe~Merve
  G{\"u}rel, Bo Li, Ce Zhang, Dawn Song, and Costas~J Spanos.
\newblock Towards efficient data valuation based on the shapley value.
\newblock In {\em The 22nd International Conference on Artificial Intelligence
  and Statistics}, pages 1167--1176. PMLR, 2019.

\bibitem{kodirov2017semantic}
Elyor Kodirov, Tao Xiang, and Shaogang Gong.
\newblock Semantic autoencoder for zero-shot learning.
\newblock In {\em Proceedings of the IEEE conference on computer vision and
  pattern recognition}, pages 3174--3183, 2017.

\bibitem{hmdb}
Hildegard Kuehne, Hueihan Jhuang, Est{\'\i}baliz Garrote, Tomaso Poggio, and
  Thomas Serre.
\newblock Hmdb: a large video database for human motion recognition.
\newblock In {\em 2011 International Conference on Computer Vision}, pages
  2556--2563. IEEE, 2011.

\bibitem{lampert2009learning}
Christoph~H Lampert, Hannes Nickisch, and Stefan Harmeling.
\newblock Learning to detect unseen object classes by between-class attribute
  transfer.
\newblock In {\em 2009 IEEE Conference on Computer Vision and Pattern
  Recognition}, pages 951--958. IEEE, 2009.

\bibitem{lampert2013attribute}
Christoph~H Lampert, Hannes Nickisch, and Stefan Harmeling.
\newblock Attribute-based classification for zero-shot visual object
  categorization.
\newblock {\em IEEE transactions on pattern analysis and machine intelligence},
  36(3):453--465, 2013.

\bibitem{rest}
Chung-Ching Lin, Kevin Lin, Lijuan Wang, Zicheng Liu, and Linjie Li.
\newblock Cross-modal representation learning for zero-shot action recognition.
\newblock In {\em Proceedings of the IEEE/CVF Conference on Computer Vision and
  Pattern Recognition}, pages 19978--19988, 2022.

\bibitem{nerenet}
Jingren Liu, Haoyue Bai, Haofeng Zhang, and Li Liu.
\newblock Near-real feature generative network for generalized zero-shot
  learning.
\newblock In {\em 2021 IEEE International Conference on Multimedia and Expo
  (ICME)}, pages 1--6. IEEE, 2021.

\bibitem{OD}
Devraj Mandal, Sanath Narayan, Sai~Kumar Dwivedi, Vikram Gupta, Shuaib Ahmed,
  Fahad~Shahbaz Khan, and Ling Shao.
\newblock Out-of-distribution detection for generalized zero-shot action
  recognition.
\newblock In {\em Proceedings of the IEEE Conference on Computer Vision and
  Pattern Recognition}, pages 9985--9993, 2019.

\bibitem{word2vec}
Tomas Mikolov, Ilya Sutskever, Kai Chen, Greg~S Corrado, and Jeff Dean.
\newblock Distributed representations of words and phrases and their
  compositionality.
\newblock In {\em Advances in neural information processing systems}, pages
  3111--3119, 2013.

\bibitem{syn}
Ashish Mishra, Anubha Pandey, and Hema~A Murthy.
\newblock Zero-shot learning for action recognition using synthesized features.
\newblock {\em Neurocomputing}, 390:117--130, 2020.

\bibitem{GGM2018}
Ashish Mishra, Vinay~Kumar Verma, M~Shiva~Krishna Reddy, S Arulkumar, Piyush
  Rai, and Anurag Mittal.
\newblock A generative approach to zero-shot and few-shot action recognition.
\newblock In {\em 2018 IEEE Winter Conference on Applications of Computer
  Vision (WACV)}, pages 372--380. IEEE, 2018.

\bibitem{olympics}
Juan~Carlos Niebles, Chih-Wei Chen, and Li Fei-Fei.
\newblock Modeling temporal structure of decomposable motion segments for
  activity classification.
\newblock In {\em European conference on computer vision}, pages 392--405.
  Springer, 2010.

\bibitem{nilsback2008automated}
Maria-Elena Nilsback and Andrew Zisserman.
\newblock Automated flower classification over a large number of classes.
\newblock In {\em 2008 Sixth Indian Conference on Computer Vision, Graphics \&
  Image Processing}, pages 722--729. IEEE, 2008.

\bibitem{palatucci2009zero}
Mark Palatucci, Dean Pomerleau, Geoffrey~E Hinton, and Tom~M Mitchell.
\newblock Zero-shot learning with semantic output codes.
\newblock {\em Advances in neural information processing systems}, 22, 2009.

\bibitem{jigsaw}
Yijun Qian, Lijun Yu, Wenhe Liu, and Alexander~G Hauptmann.
\newblock Rethinking zero-shot action recognition: Learning from latent atomic
  actions.
\newblock In {\em European Conference on Computer Vision}, pages 104--120.
  Springer, 2022.

\bibitem{qin2017zero}
Jie Qin, Li Liu, Ling Shao, Fumin Shen, Bingbing Ni, Jiaxin Chen, and Yunhong
  Wang.
\newblock Zero-shot action recognition with error-correcting output codes.
\newblock In {\em Proceedings of the IEEE Conference on Computer Vision and
  Pattern Recognition}, pages 2833--2842, 2017.

\bibitem{rohrbach12eccv}
Marcus Rohrbach, Michaela Regneri, Mykhaylo Andriluka, Sikandar Amin, Manfred
  Pinkal, and Bernt Schiele.
\newblock {Script data for attribute-based recognition of composite
  activities}.
\newblock In {\em ECCV}, 2012.

\bibitem{schonfeld2019generalized}
Edgar Schonfeld, Sayna Ebrahimi, Samarth Sinha, Trevor Darrell, and Zeynep
  Akata.
\newblock Generalized zero-and few-shot learning via aligned variational
  autoencoders.
\newblock In {\em Proceedings of the IEEE/CVF Conference on Computer Vision and
  Pattern Recognition}, pages 8247--8255, 2019.

\bibitem{trpo}
John Schulman, Sergey Levine, Pieter Abbeel, Michael Jordan, and Philipp
  Moritz.
\newblock Trust region policy optimization.
\newblock In {\em International conference on machine learning}, pages
  1889--1897. PMLR, 2015.

\bibitem{ppo}
John Schulman, Filip Wolski, Prafulla Dhariwal, Alec Radford, and Oleg Klimov.
\newblock Proximal policy optimization algorithms.
\newblock {\em arXiv preprint arXiv:1707.06347}, 2017.

\bibitem{vgg}
Karen Simonyan and Andrew Zisserman.
\newblock Very deep convolutional networks for large-scale image recognition.
\newblock {\em arXiv preprint arXiv:1409.1556}, 2014.

\bibitem{socher2013zero}
Richard Socher, Milind Ganjoo, Christopher~D Manning, and Andrew Ng.
\newblock Zero-shot learning through cross-modal transfer.
\newblock {\em Advances in neural information processing systems}, 26, 2013.

\bibitem{ucf101}
Khurram Soomro, Amir~Roshan Zamir, and Mubarak Shah.
\newblock Ucf101: A dataset of 101 human actions classes from videos in the
  wild.
\newblock {\em arXiv preprint arXiv:1212.0402}, 2012.

\bibitem{reinforce}
Richard~S Sutton, David McAllester, Satinder Singh, and Yishay Mansour.
\newblock Policy gradient methods for reinforcement learning with function
  approximation.
\newblock {\em Advances in neural information processing systems}, 12, 1999.

\bibitem{vaswani2017attention}
Ashish Vaswani, Noam Shazeer, Niki Parmar, Jakob Uszkoreit, Llion Jones,
  Aidan~N Gomez, {\L}ukasz Kaiser, and Illia Polosukhin.
\newblock Attention is all you need.
\newblock {\em Advances in neural information processing systems}, 30, 2017.

\bibitem{verma2018generalized}
Vinay~Kumar Verma, Gundeep Arora, Ashish Mishra, and Piyush Rai.
\newblock Generalized zero-shot learning via synthesized examples.
\newblock In {\em Proceedings of the IEEE conference on computer vision and
  pattern recognition}, pages 4281--4289, 2018.

\bibitem{cub}
Catherine Wah, Steve Branson, Peter Welinder, Pietro Perona, and Serge
  Belongie.
\newblock The caltech-ucsd birds-200-2011 dataset.
\newblock 2011.

\bibitem{wu2022davinz}
Zhaoxuan Wu, Yao Shu, and Bryan Kian~Hsiang Low.
\newblock Davinz: Data valuation using deep neural networks at initialization.
\newblock In {\em International Conference on Machine Learning}, pages
  24150--24176. PMLR, 2022.

\bibitem{xian2018zero}
Yongqin Xian, Christoph~H Lampert, Bernt Schiele, and Zeynep Akata.
\newblock Zero-shot learning—a comprehensive evaluation of the good, the bad
  and the ugly.
\newblock {\em IEEE transactions on pattern analysis and machine intelligence},
  41(9):2251--2265, 2018.

\bibitem{clswgan}
Yongqin Xian, Tobias Lorenz, Bernt Schiele, and Zeynep Akata.
\newblock Feature generating networks for zero-shot learning.
\newblock In {\em Proceedings of the IEEE conference on computer vision and
  pattern recognition}, pages 5542--5551, 2018.

\bibitem{xian2017zero}
Yongqin Xian, Bernt Schiele, and Zeynep Akata.
\newblock Zero-shot learning-the good, the bad and the ugly.
\newblock In {\em Proceedings of the IEEE Conference on Computer Vision and
  Pattern Recognition}, pages 4582--4591, 2017.

\bibitem{fvaegan}
Yongqin Xian, Saurabh Sharma, Bernt Schiele, and Zeynep Akata.
\newblock f-vaegan-d2: A feature generating framework for any-shot learning.
\newblock In {\em Proceedings of the IEEE/CVF conference on computer vision and
  pattern recognition}, pages 10275--10284, 2019.

\bibitem{sun}
Jianxiong Xiao, James Hays, Krista~A Ehinger, Aude Oliva, and Antonio Torralba.
\newblock Sun database: Large-scale scene recognition from abbey to zoo.
\newblock In {\em 2010 IEEE computer society conference on computer vision and
  pattern recognition}, pages 3485--3492. IEEE, 2010.

\bibitem{xu2017transductive}
Xun Xu, Timothy Hospedales, and Shaogang Gong.
\newblock Transductive zero-shot action recognition by word-vector embedding.
\newblock {\em International Journal of Computer Vision}, 123(3):309--333,
  2017.

\bibitem{xu2016multi}
Xun Xu, Timothy~M Hospedales, and Shaogang Gong.
\newblock Multi-task zero-shot action recognition with prioritised data
  augmentation.
\newblock In {\em European Conference on Computer Vision}, pages 343--359.
  Springer, 2016.

\bibitem{cmcnet}
Fu-En Yang, Yuan-Hao Lee, Chia-Ching Lin, and Yu-Chiang~Frank Wang.
\newblock Semantics-guided intra-category knowledge transfer for generalized
  zero-shot learning.
\newblock {\em International Journal of Computer Vision}, pages 1--15, 2023.

\bibitem{zerogen}
Jiacheng Ye, Jiahui Gao, Qintong Li, Hang Xu, Jiangtao Feng, Zhiyong Wu, Tao
  Yu, and Lingpeng Kong.
\newblock Zerogen: Efficient zero-shot learning via dataset generation.
\newblock {\em arXiv preprint arXiv:2202.07922}, 2022.

\bibitem{ye2020synthetic}
Jiarong Ye, Yuan Xue, L~Rodney Long, Sameer Antani, Zhiyun Xue, Keith~C Cheng,
  and Xiaolei Huang.
\newblock Synthetic sample selection via reinforcement learning.
\newblock In {\em Medical Image Computing and Computer Assisted
  Intervention--MICCAI 2020: 23rd International Conference, Lima, Peru, October
  4--8, 2020, Proceedings, Part I 23}, pages 53--63. Springer, 2020.

\bibitem{ye2017zero}
Meng Ye and Yuhong Guo.
\newblock Zero-shot classification with discriminative semantic representation
  learning.
\newblock In {\em Proceedings of the IEEE conference on computer vision and
  pattern recognition}, pages 7140--7148, 2017.

\bibitem{yoon2020data}
Jinsung Yoon, Sercan Arik, and Tomas Pfister.
\newblock Data valuation using reinforcement learning.
\newblock In {\em International Conference on Machine Learning}, pages
  10842--10851. PMLR, 2020.

\bibitem{yu2020episode}
Yunlong Yu, Zhong Ji, Jungong Han, and Zhongfei Zhang.
\newblock Episode-based prototype generating network for zero-shot learning.
\newblock In {\em Proceedings of the IEEE/CVF conference on computer vision and
  pattern recognition}, pages 14035--14044, 2020.

\bibitem{daa}
Xiaojie Zhao, Yuming Shen, Shidong Wang, and Haofeng Zhang.
\newblock Generating diverse augmented attributes for generalized zero shot
  learning.
\newblock {\em Pattern Recognition Letters}, 2023.

\end{thebibliography}
}

\end{document}